\documentclass[preprint,12pt]{elsarticle}

\usepackage{a4wide}

\usepackage{amssymb}
\usepackage{amsmath}
\usepackage{physics}
\usepackage{booktabs} 
\usepackage{multirow}
\usepackage{makecell}
\usepackage{tabularx}
\usepackage{calc}
\usepackage{url}          
\usepackage{hyperref}     
\usepackage{xurl}         

\hypersetup{
    colorlinks=true,      
    linkcolor=blue,       
    citecolor=blue,       
    urlcolor=blue,        
    breaklinks=true       
}

\begin{document}

\begin{frontmatter}



\title{Generative Modeling of Clinical Time Series via Latent Stochastic Differential Equations} 

\author[label1]{Muhammad Aslanimoghanloo} 
\author[label1]{Ahmed ElGazzar} 
\author[label1]{Marcel van Gerven} 

\affiliation[label1]{organization={Department of Machine Learning and Neural Computing, Donders Institute for Brain, Cognition and Behaviour, Radboud University},
            city={Nijmegen},
            postcode={6525 GD}, 
            country={the Netherlands}}

\begin{abstract}
Clinical time series data from electronic health records and medical registries offer unprecedented opportunities to understand patient trajectories and inform medical decision-making. However, leveraging such data presents significant challenges due to irregular sampling, complex latent physiology, and inherent uncertainties in both measurements and disease progression.  To address these challenges, we propose a generative modeling framework based on latent neural stochastic differential equations (SDEs) that views clinical time series as discrete-time partial observations of an underlying controlled stochastic dynamical system. Our approach models latent dynamics via neural SDEs with modality-dependent emission models, while performing state estimation and parameter learning through variational inference. This formulation naturally handles irregularly sampled observations, learns complex non-linear interactions, and captures the stochasticity of disease progression and measurement noise within a unified scalable probabilistic framework. We validate the framework on two complementary tasks: (i) individual treatment effect estimation using a simulated pharmacokinetic-pharmacodynamic (PKPD) model of lung cancer, and (ii) probabilistic forecasting of physiological signals using real-world intensive care unit (ICU) data from 12,000 patients. Results show that our framework outperforms ordinary differential equation and long short-term memory baseline models in accuracy and uncertainty estimation. These results highlight its potential for enabling precise, uncertainty-aware predictions to support clinical decision-making.
\end{abstract}

\end{frontmatter}

\section{Introduction}
\label{intro}
Predicting patient trajectories is critical for enabling timely interventions, better understanding of disease progression, and developing personalized medicine~\citep{dixon2024unveiling}. For instance, early detection of sepsis has been shown to significantly reduce the risk of organ failure and mortality~\citep{markham2025patient}. 
This potential is increasingly becoming feasible due to the rapid growth of available healthcare data like electronic health records (EHRs)~\citep{yin2023prediction}. 
A defining feature of healthcare data are their temporal nature, reflecting the dynamic evolution of patient conditions over time. For instance, a significant rise in temperature and heart rate over several hours might indicate an infection~\citep{patharkar2024predictive}. These temporal patterns highlight the need for time series models specifically tailored to the complexities of clinical data.
However, healthcare time series have unique characteristics such as missing values, irregular sampling, aleatoric uncertainty, and patient-specific variability, that make modeling them particularly challenging~\citep{noy2023time, liu2016learning}.

Traditional time series models, such as autoregressive moving average (ARIMA) models, have been applied to healthcare data but often struggle with its complexity and irregularity~\citep{morid2023time}. 
The deep learning era brought significant advances, with recurrent neural networks (RNNs), long short-term memory networks (LSTMs), and gated recurrent units (GRUs) becoming widely adopted for modeling clinical longitudinal data~\citep{cascarano2023machine}. Notable examples include Doctor AI, which used an RNN to predict future diagnoses and medications from a patient’s sequential medical history, and RETAIN, which added an attention mechanism for interpretability in predicting heart failure onset~\citep{choi2016doctor,choi2016retain}. These studies revealed that neural networks could automatically learn representations of high-dimensional EHR sequences, outperforming traditional models that relied on manually selected variables. However, handling irregularity and missing values still remains a challenge for these models.

Neural ordinary differential equations (neural ODEs) marked a paradigm shift in time series modeling by enabling continuous-time modeling~\citep{chen2018neural}. Unlike traditional discrete-time approaches that process data at fixed intervals, neural ODEs model hidden state evolution as solutions to continuous differential equations parameterized by neural networks, enabling natural handling of irregularly-sampled observations~\citep{rubanova2019latent}. This continuous-time formulation has proven particularly valuable in healthcare settings, where neural ODE-based approaches have been successfully applied to model counterfactual treatment outcomes, enabling clinicians to predict how patients might respond to alternative treatment strategies~\citep{seedat2022continuous}. Furthermore, researchers have leveraged ODE discovery methods to infer interpretable differential equations that govern longitudinal treatment effects, providing both predictive capability and mechanistic insights into patient-specific treatment responses. Moreover, neural ODE frameworks allow for easier integration of domain knowledge, enhancing their prediction performance, interpretability, and explainability~\citep{kacprzyk2024ode}. The flexibility of neural ODEs to incorporate external control signals, such as medication dosages or intervention schedules, through neural controlled differential equations has further expanded their applicability to complex clinical scenarios where treatment decisions must be optimized over time~\citep{kidger2020neural}.

Although neural ODEs are powerful, continuous-time, and efficient, they are deterministic and lack the ability to model stochasticity and provide uncertainty quantification. However, many real-world systems, particularly in biology and healthcare, are inherently stochastic and require  uncertainty estimates for reliable decision-making. To address these limitations, we propose a latent neural stochastic differential equation (latent SDE) framework that views clinical time series as discrete-time partial observations of an underlying controlled stochastic dynamical system. Our approach models latent dynamics via neural stochastic differential equations (neural SDEs) with a modality-dependent emission model, while performing state estimation and parameter learning through variational inference. Neural SDEs extend neural ODEs by incorporating two main components: the drift part and the diffusion part. The drift term represents the deterministic component of the system's evolution, while the diffusion term models the stochastic part in the system dynamics. Similar to neural ODEs, these terms are typically parameterized by neural networks, enabling the model to learn intricate stochastic dynamics directly from the data~\citep{kidger2022neural,tzen2019neural,li2020scalable}. Moreover, neural SDEs enable uncertainty quantification by producing a distribution over possible future trajectories, rather than a single deterministic forecast, providing both expected outcomes and associated uncertainties. This capability is crucial for informed clinical decision-making and risk assessment~\citep{kong2020sde}. We validate our proposed latent SDE framework on two tasks: treatment effect estimation using cancer simulation data, and physiological time series forecasting using ICU data. These experiments demonstrate the framework’s flexibility and effectiveness in clinical time series modeling, highlighting its potential to advance data-driven healthcare.

\section{Related works}
In this section, we review related work on clinical time series modeling in continuous-time settings, with particular focus on neural differential equation models and uncertainty-aware methods.

\subsection{Neural ODEs/SDEs}
Neural ordinary differential equations can be viewed as infinitely deep residual neural networks in which the hidden state evolution is parameterized by a continuous-time ordinary differential equation~\citep{chen2018neural}. This allows the model to operate on irregularly-sampled time series, use adaptive computation, and produce memory-efficient gradients via the adjoint sensitivity method.
Neural controlled differential equations extend this framework by allowing the hidden state evolution to be driven by a control signal rather than relying solely on an initial condition~\citep{kidger2020neural}.
Generative models for time series have been developed by integrating neural ODEs into a variational autoencoder (VAEs) framework. The key idea is to represent the latent dynamics of the time series using a continuous-time ordinary differential equation, which proves especially effective when dealing with partial, noisy observations of underlying system states~\citep{rubanova2019latent,garsdal2022generative}.
Latent SDEs further generalize the latent neural ODE framework by modeling latent dynamics as stochastic processes. A variational inference approach based on Girsanov’s theorem enables the construction of an evidence lower bound (ELBO) and supports training with black-box SDE solvers~\citep{tzen2019neural}. To address the challenge of gradient computation in this setting, the stochastic adjoint sensitivity method was developed, allowing for scalable training via stochastic variational inference~\citep{li2020scalable}.

\subsection{Clinical time series modeling in continuous-time}
Several studies have applied continuous-time models to estimate treatment effects over time. \citet{kacprzyk2024ode} utilized ODE discovery methods for estimating longitudinal heterogeneous treatment effects. Unlike existing neural network-based models that often act as black-box predictors, the proposed approach learns interpretable, closed-form ODEs that govern individual patient dynamics under different treatment regimens.
\citet{laurie2023explainable} developed an explainable neural ODE framework that models tumor growth dynamics and predicts overall survival in cancer patients. This approach integrates clinical covariates and tumor dynamics into a unified model to predict patient survival outcomes. To enhance interpretability, they introduce mechanisms for feature attribution and trajectory analysis, allowing clinicians to understand which factors most influence survival predictions. Similarly \citet{lu2021neural} used neural ODEs for pharmacokinetics modeling and demonstrated that it offers a flexible and continuous-time alternative to traditional machine learning models, which often struggle with generalization to unseen dosing regimens. Additionally, domain knowledge has been integrated into neural ODE-based models by integrating a system of expert-designed ODEs with machine-learned Neural ODEs to fully describe the dynamics of the system and to link the expert and latent variables to observable quantities~\citep{qian2021integrating}.
However, these ODE-based methods lack uncertainty quantification and provide only point estimates, limiting their applicability in high-stakes clinical settings where understanding prediction confidence is essential.

\subsection{Uncertainty-aware modeling}
Recognizing the limitation of deterministic models, researchers have begun combining Bayesian inference with neural differential equations to enable probabilistic forecasting for treatment effect estimation. These methods combine the flexibility of differential equations with Bayesian inference to capture uncertainty in both model parameters and predictions~\citep{hess2023bayesian,de2022predicting}. However, they are mainly based on ordinary differential equations which have limited capability in modeling stochastic processes like patient trajectories. 

\subsection{Our contribution}
Building on these foundations, we propose a framework that integrates neural stochastic differential equations with a variational inference-based encoder–decoder architecture for modeling clinical time series in latent space. This approach enables continuous-time modeling of patient trajectories while capturing predictive uncertainty, thereby supporting more robust and data-driven decision-making in healthcare.

\section{Problem formulation}
Let $t=[0,\tau]$ be the observation time window during which patient status is observed through clinical measurements $y(t) \in \mathbb{R}^{d_y}$ (e.g., heart rate, tumor volume) which may be recorded at irregular time points. 
Each patient $n \in \{1,\ldots,N\}$, may receive treatments $$U^n= \{u^n(t) \in \mathbb{R}^{d_u} \mid t \in [0,\tau] \}$$ administered at potentially irregular times which may differ from measurement time points. Additionally, each patient has baseline covariates $z \in \mathcal{Z}$ (e.g., age, BMI, sex, and disease subtype) which may influence both disease progression and treatment response. 
Accordingly, we represent a group of $N$ patients as a set of continuous-time observational profiles: 
\begin{align}
    \mathcal{P}= \{P^n=(y^n(t), z^n, u^n(t)) \}_{n=1}^N 
\end{align}
where each $P^n$ denotes the observational record of the $n$-th patient, encompassing longitudinal outcomes, baseline covariates, and treatment history over time, collectively referred to as patient trajectories.

Given the historical trajectory $P^n$ of patient $n$ up to time $\tau$, our goal is to estimate their potential outcomes $y^n(t; u_{[\tau, \tau']})$ for an arbitrary future sequence of treatments $U^n(t)$ over a future time window, $t \in [\tau, \tau']$, $
\tau' > \tau$. 

To account for the complex and uncertain nature of clinical data, we do not restrict our model to point estimates. Instead, we aim to learn the full conditional distribution over future outcomes:
\begin{align}
f: P^n \mapsto \mathbb{P}\left(y^n(t) \mid P^n, u^n_{[\tau, \tau']}\right)  \text{ for } t \in [\tau, \tau']\,,
\end{align}
allowing us to quantify predictive uncertainty, which is essential for reliable and risk-aware clinical decision-making.

We assume that observations arise from an underlying latent state process $x^n(t) \in \mathbb{R}^{d_x}$, which evolves continuously over time, driven by the patient’s treatment history and baseline covariates. The observations $y^n(t)$ are treated as noisy emissions from this latent trajectory. This formulation naturally handles irregular sampling and enables continuous-time interpolation and extrapolation.

\section{Latent stochastic differential equation framework}

We propose a probabilistic generative modeling framework that interprets patient time series data as partial observations of an underlying controlled stochastic dynamical system evolving in continuous time. 
Specifically, we model the latent dynamics using neural stochastic differential equations, where both the drift and diffusion terms are parameterized by neural networks and conditioned on external inputs such as treatments. 
This formulation enables the model to capture temporal uncertainty and variability in disease progression while flexibly handling irregularly-sampled measurements and interventions.

\subsection{Architecture}
\begin{figure}[ht]
     \centering
 \includegraphics[width=1.0\textwidth]{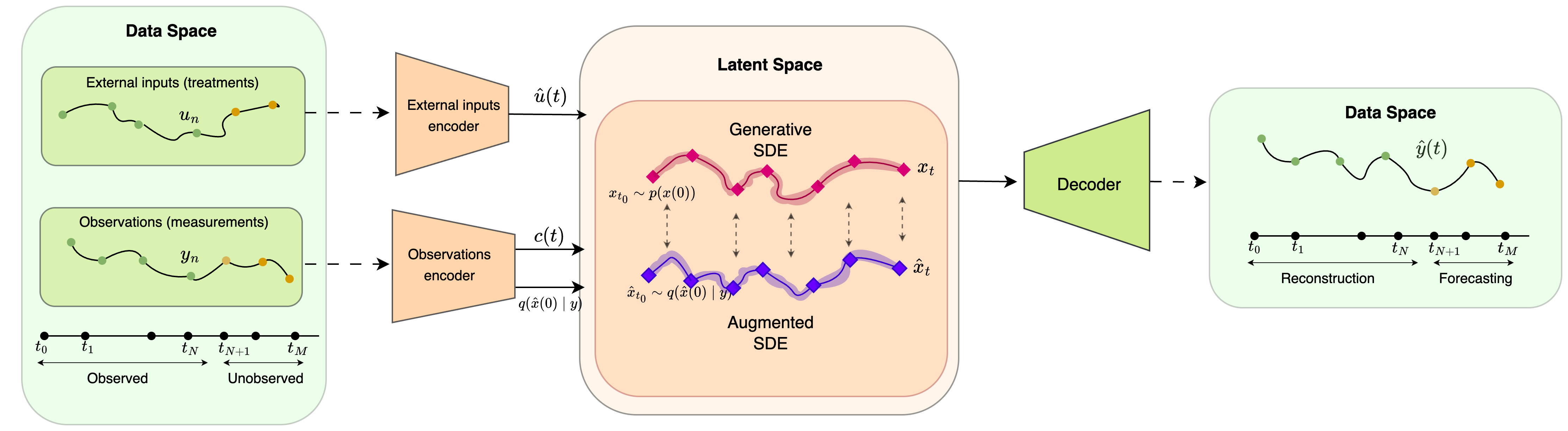}
     \caption{Architecture of the latent SDE framework for time series modeling with external inputs. The model consists of three main components: (1) Data encoders that transform external inputs and observations into latent representations; (2) A latent space containing both a generative SDE and an augmented SDE that evolve the hidden state conditioned on the encoded inputs; and (3) A decoder that reconstructs the output in the data space. The framework enables both reconstruction of observed data and forecasting of future values by leveraging the continuous-time dynamics learned through the coupled SDE system.}
     \label{fig:lsde_arch}
 \end{figure}

An overview of the proposed framework is shown in Figure \ref{fig:lsde_arch}. The architecture follows a variational encoder–decoder structure, with a latent neural SDE acting as the generative model. This latent representation allows the model to operate in a lower-dimensional space while capturing complex, nonlinear, and uncertain clinical dynamics. 
We employ two separate encoders: one for observations, which maps discrete-time measurements $Y^n$ and covariates $Z^n$ into the latent space, and another for treatments $A^n$, which encodes the sequence of interventions into a continuous-time control signal. This continuous-time representation of the treatments enables the generative model to flexibly incorporate interventions at arbitrary time points during simulation or forecasting.

We treat the latent state evolution as a dynamical system governed by a stochastic differential equation $x:[0,\tau] \rightarrow \mathbb{R}^{d_x}$. The latent state is indirectly observed through discrete-time measurements ( e.g., clinical measurements), denoted by $\mathcal{Y}=\{\mathrm{y}_i\}_{i=1}^T$ where each $\mathrm{y}_i \in \mathbb{R}^{d_y}$ corresponds to an observation at time $t_i$ and $T$ is the last time point in measurements. The system may also be influenced by external inputs — in this case, treatments — represented as $\mathcal{V}=\{\mathrm{v}_i\}_{i=1}^T$, with each input $\mathrm{v}_i \in \mathbb{R}^{d_v}$ possibly occurring at different times than the observations.
Once the system dynamics are modeled in the latent space using neural SDEs, an observation decoder is used to project the latent trajectories back into the observational space. The resulting latent SDE model is equivalent to a continuous-time stochastic state space model which enables simulation of future outcomes under new treatment plans and provides uncertainty estimates through stochastic sampling.

\subsection{Generative model}
We model latent states dynamics $x$ as a continuous-time stochastic process, specified as the solution to the following It\^o SDE: 
\begin{equation}
    \dd x(t)=\mu_\theta(x(t),u(t)) \dd t + \sigma_\theta(x(t),u(t)) \dd W(t) 
    \label{ito_SDE}
\end{equation}
where $x(0)$ is drawn from a learned distribution, $\mathcal{P}_0$, $\mu_\theta$ and $\sigma_\theta$ are drift and diffusion terms parameterized by neural networks with learnable parameters ($\theta$), $u(t)$ is the encoded control input (e.g., treatment signal), and $W(t)$ is a standard Wiener process.

This generative SDE induces a probability measure over the space of continuous trajectories, thereby implicitly defining the conditional distribution $p(x \mid \mathcal{V})$. By solving the SDE multiple times with different realizations of the initial condition $x(0) \sim \mathcal{P}_0$ and independent Wiener process paths $w(t)$, we can obtain samples from this distribution, each corresponding to a possible latent trajectory given the external inputs $\mathcal{V}$:
\begin{align}
    x(t)= \int^t_0 \mu_\theta(x(s), u(s)) \dd s +\int_0^t \sigma_\theta(x(s),u(s))\dd w(s) \,.
    \label{sde_integral}
\end{align}

\subsubsection{Variational inference}
The exact posterior distribution $p(x \mid \mathcal{D})$, given data $\mathcal{D}$, is intractable to compute since computing the exact posterior involves integrating over all possible realizations of $w(t)$, which is computationally infeasible. To address this, we employ variational inference, which offers an efficient and scalable alternative. Specifically, we approximate the true posterior $p(x \mid \mathcal{D})$ using a tractable family of parameterized distributions $q_\phi(x \mid D)$, and optimize the parameters $\phi$ by maximizing a variational lower bound. 

We define the approximate posterior distribution $q_\phi(x \mid \mathcal{D})$ via an augmented SDE~\citep{li2020scalable}. 
An augmented SDE is an auxiliary SDE whose drift term is conditioned on the observed data, while its diffusion term $\sigma_\theta$ is shared with the generative SDE, ensuring that the KL divergence between the two stochastic processes remains tractable via Girsanov's theorem~\citep{girsanov1960transforming}. 
Consequently, the approximate posterior path $\Tilde{x}:[0,\tau] \rightarrow \mathbb{R}^{d_x}$ is modeled as a solution to the following augmented SDE (omitting explicit time indices for clarity): 
\begin{align}
    \dd \Tilde{x}=\nu_\phi(\Tilde{x},u,c)\dd t +\sigma_\theta(\Tilde{x},u)\dd W
    \label{aug_sde}
\end{align}
where $\Tilde{x}(0)$ is sampled from a data-dependent distribution $\mathcal{Q}_0=\mathcal{N}(\alpha_\phi(\mathrm{y}_{1:c}), \beta_\phi(\mathrm{y}_{1:c}))$ with $\alpha_\phi$ and $\beta_\phi$ being neural networks that encode the initial observations $\mathrm{y}_{1:c}=\{\mathrm{y}_1,\ldots, \mathrm{y}_c\}$. To avoid over-parameterizing the initial condition and thereby diminishing the role of the learned dynamics, we restrict this encoding to only the first $c$ observations (with $t_c \leq t_n$) rather than the entire observation sequence~\citep{elgazzar2024generative}.
The function $c(t)$ denotes a continuous-time context signal constructed from the observed data, providing additional conditioning for the variational drift term $v_\phi$. 

Consequently, the Kullback–Leibler (KL) divergence between the path measures of the augmented ($\mathcal{Q}_\tau$) and generative  ($\mathcal{P}_\tau$) SDEs is given by:
\begin{equation}
D_{\mathrm{KL}}(\mathcal{Q}_\tau \parallel \mathcal{P}_\tau) = \mathbb{E}_{\tilde{x}} \left[ \int_0^\tau \frac{1}{2} \left\| \Delta(\tilde{x}, u, c) \right\|^2 \, dt \right]
\end{equation}
where \( \Delta(\tilde{x}, u, c) = \sigma_\theta(\tilde{x}, u)^{-1} \left( \nu_\phi(\tilde{x}, u, c) - \mu_\theta(\tilde{x}, u) \right) \) denotes the likelihood ratio drift, also known as the Radon--Nikodym derivative
of the drift adjustment under Girsanov's theorem~\citep{kallenberg1997foundations}. 

By minimizing this divergence, we ensure that generative SDE closely aligns with the augmented SDE in distribution. Simultaneously, we require that the latent trajectories produced by the augmented SDE, when decoded back into the data space, closely match the observed data. 
Consequently, we can define an evidence lower bound (ELBO) on the conditional marginal likelihood of the observations as follows: 
\begin{align}
\log p(\mathcal{Y}, \mathcal{V}) \geq
\mathbb{E}_{\tilde{x}} \left[
\sum_{i=1}^{T} \log p(y_i \mid \tilde{x}_i) 
\right]
- D_{\mathrm{KL}}(\mathcal{Q}_0 \| \mathcal{P}_0) - D_{\mathrm{KL}}(\mathcal{Q}_\tau \| \mathcal{P}_\tau) \, .
\end{align}
The first term on the right-hand side represents the expected log-likelihood of the observations given the latent states $\tilde{x}_i$ sampled from the augmented SDE. This term encourages the model to reconstruct the data accurately. The second and third are KL divergence terms that regularize the variational distribution: $D_{\mathrm{KL}}(\mathcal{Q}_0 \parallel \mathcal{P}_0)$ between the augmented and generative SDEs initial condition distributions, and $D_\mathrm{KL}(\mathcal{Q}_\tau \parallel \mathcal{P}_\tau)$ measures the divergence between the entire approximate posterior and prior trajectory distributions. 
The model parameters  $\theta$ and the variational parameters $\phi$ are jointly optimized by maximizing the ELBO using stochastic gradient descent. Once trained, the generative component of the model can be used independently for downstream tasks such as forecasting future observations, reconstructing missing values, or simulating counterfactual scenarios under alternative treatment plans.

\section{Experiments}
We evaluated our proposed framework on two datasets: (i) a synthetic dataset generated using a pharmacokinetic-pharmacodynamic (PKPD) model of lung cancer tumor growth, and (ii) a real-world dataset from hospital ICUs. In both cases, the objective was to learn the underlying system dynamics and forecast patients' future trajectories. 
This dual evaluation strategy allows us to assess performance under controlled conditions with known ground truth (synthetic data) and validate practical applicability in realistic clinical settings (real-world data).

We compared our latent SDE model against two established baselines representing different modeling approaches: (i) the latent ODE, a deterministic continuous-time model~\citep{rubanova2019latent}, and (ii) the latent LSTM, representing the standard discrete-time approach commonly used for clinical time series. This comparison helps us understand the contributions of stochastic dynamics (SDE vs. ODE) and time formulation (continuous vs. discrete) to model performance. Both baselines use the same encoder-decoder architecture as our proposed model to ensure fair comparison, isolating the impact of the latent dynamics modeling approach.

For each dataset, we employed a temporal segmentation, where models used an initial observation window to predict subsequent time points. This setup mimics realistic clinical forecasting scenarios where predictions must be made based on available patient history. Furthermore, we systematically evaluated model robustness under three challenging conditions commonly encountered in clinical practice: irregular measurement timing, sparse observational data (20-80\% missing values), and varying levels of process noise. These robustness studies assess whether the observed performance differences persist under realistic data degradation conditions.

\subsection{Data}

\subsubsection{Synthetic dataset}
To carry out controlled experiments with known ground truth, we generated a synthetic dataset based on a pharmacokinetic-pharmacodynamic (PKPD) model of lung cancer tumor growth. This model is commonly used to simulate the effects of chemotherapy and radiotherapy on tumor dynamics~\citep{kacprzyk2024ode}. Our objective was to predict treatment outcomes from partial observations of the underlying physiological states. 

To create a more realistic clinical scenario, we extended the basic tumor growth model to include two additional state variables: immune system response and patient health status, each governed by separate differential equations that interact with tumor dynamics and treatment effects. This extension captures the complex interactions between tumor dynamics, immune function, and patient well-being that occur in actual clinical settings. The immune system responds to both the tumor burden and treatment effects, while patient health status reflects the combined impact of disease progression, immune function, and treatment toxicity.

To reflect clinically realistic measurement processes, we modeled indirect observations of the system states rather than direct access to underlying variables. Accordingly, tumor volume is indirectly observed through noisy cancer cell counts, which we model using a Poisson distribution to reflect the discrete, count-based nature of cell counts. Patient's health status is observed through Eastern Cooperative Oncology Group (ECOG) performance status scores, a standardized 6-point ordinal scale ranging from 0 (fully active) to 5 (deceased) commonly used in oncology to assess patients' functional status~\citep{oken1982ecog}.

To capture inter-patient variability, we incorporated five baseline covariates: age, gender, weight, height, and tumor type. These covariates modulate the pharmacokinetic-pharmacodynamic model parameters, creating patient-specific treatment response profiles. For example, older patients exhibit reduced drug clearance rates, while body surface area (derived from height and weight) affects health status recovery. Moreover, all model parameters were drawn from appropriate probability distributions rather than fixed at deterministic values, ensuring diverse patient trajectories and realistic population-level heterogeneity~\citep{kacprzyk2024ode}.

Eventually, we introduced irregular sampling by randomly removing 20--80\% of time points from each trajectory, simulating the sparse and asynchronous measurement patterns typical in clinical practice. This approach allows us to evaluate model performance under varying levels of data availability without requiring separate datasets.
Complete model equations, parameter distributions, and simulation details are provided in \ref{appndx:synthetic_dataset}.

\subsubsection{Empirical dataset}
To evaluate our framework's performance in a real-world clinical setting, we used data from the PhysioNet Computing in Cardiology Challenge 2012, which includes longitudinal records from approximately 12,000 ICU stays lasting at least 48 hours~\citep{churpek2016value}. This dataset contains 42 variables: 6 general descriptors (age, gender, height, weight, ICU type, and record ID), 36 time series variables recorded at least once per stay (vital signs, laboratory values, and other clinical measurements), and 5 outcome-related descriptors such as survival time, in-hospital death, and length of stay.

However, in contrast to the original challenge, which focused on predicting in-hospital death, we aimed to forecast time series variables. This task is valuable for applications such as early detection of clinical deterioration and optimization of treatment plans. Among all time series variables,  we selected three vital signs for prediction: heart rate (HR), invasive mean arterial blood pressure (MAP), and body temperature (BT). These variables were chosen based on their sufficiently high sampling frequency and clinical relevance.

We prepared the empirical dataset similar to the synthetic dataset by splitting each patient's trajectory into observation and prediction windows. To address the inherent irregularity of ICU data, we retained the original time stamps without resampling to fixed intervals. Missing values were left unfilled and passed to the model as masked inputs, allowing it to directly learn from the observed timestamps and the presence or absence of measurements. This design reflects realistic clinical settings and leverages the model’s ability to handle irregular, partially observed sequences. Further details on preprocessing steps, variable selection, and dataset characteristics are provided in \ref{appndx:empirical_dataset}.

\section{Results}
In this section, we present the performance of the proposed latent SDE framework on both synthetic (PKPD) and real-world (ICU) datasets, and compare it against the baseline latent ODE and latent LSTM models. We assess each model's ability to capture underlying system dynamics and generate accurate probabilistic forecasts under different conditions, including irregular sampling, observation sparsity, and stochasticity.

\subsection{Synthetic dataset}
Table~\ref{tab:model_performance_pkpd} summarizes the performance of the latent SDE, latent ODE, and latent LSTM models on the synthetic dataset. The latent SDE model outperformed both baselines across all evaluation metrics, with particularly strong advantages in uncertainty quantification. Specifically, the model achieved substantially lower predictive entropy (PE), continuous ranked probability score (CRPS), and negative log-likelihood (NLL) compared to the deterministic latent ODE and the latent LSTM. These results demonstrate that the latent SDE effectively captures both the observed variables (performance status and cancer cell count) and the unobserved latent state (tumor volume) while providing well-calibrated uncertainty estimates.

The performance differences between models reflect their underlying formulations. The latent SDE's advantages stem from its explicit modeling of stochastic dynamics through the diffusion term, which captures the inherent randomness in biological processes and generates trajectory predictions that reflect natural variability in patient responses to treatment. This stochastic formulation enables realistic modeling of inter-patient heterogeneity and measurement noise, fundamental characteristics of clinical data. In contrast, the latent ODE model's deterministic formulation constrains it to single-valued predictions. Without a diffusion component, the ODE approach cannot represent the range of plausible patient trajectories or model stochastic fluctuations, limiting its utility for risk-aware clinical decision-making where understanding prediction confidence is essential.

The latent LSTM exhibited comparatively weaker performance, which can be attributed to several structural characteristics of recurrent variational models. Although the model incorporates probabilistic elements through variational initial conditions, the deterministic recurrent architecture causes this stochasticity to diminish over time as deterministic updates dominate trajectory evolution. This results in predicted trajectories that lose variance with increasing sequence length. Additionally, LSTM-based variational models are prone to posterior collapse, where the approximate posterior distribution converges to the prior, rendering latent variables less informative for capturing patient-specific variability. The sequential processing inherent to LSTMs also presents challenges for modeling long-term dependencies, as vanishing gradients can limit the effective temporal range despite gating mechanisms designed to mitigate this issue.
\begin{table}[ht]
\centering
\caption{Performance comparison of latent SDE, latent ODE, and latent LSTM models on the synthetic PKPD dataset. Lower values are better for RMSE, PE, CRPS, and NLL; higher values are better for accuracy.}
\label{tab:model_performance_pkpd}
\setlength{\tabcolsep}{8pt}
\begin{small}
\begin{tabular}{l l c c c}
\toprule
 \textbf{Target Variable} &  \textbf{Metric} &  \makecell{\textbf{Latent} \\ \textbf{SDE}} &  \makecell{\textbf{Latent} \\ \textbf{ODE}} & \makecell{\textbf{Latent} \\ \textbf{LSTM}} \\
\midrule
\multirow{2}{*}{Performance status} & Acc & $\textbf{\hspace{\widthof{1}}0.56} \pm \textbf{0.08}$ & $\hspace{\widthof{1}}0.47\pm 0.11$ & $\hspace{\widthof{1}}0.44\pm 0.07$ \\
                    & PE & $\textbf{\hspace{\widthof{1}}4.56} \pm \textbf{0.69}$ & $\hspace{\widthof{1}}6.43 \pm 0.47$ & $\hspace{\widthof{1}}8.56 \pm 0.21$\\
\midrule
\multirow{2}{*}{\makecell{ Tumor volume \\ (unobserved)}} & RMSE & $\textbf{\hspace{\widthof{1}}4.27} \pm \textbf{0.06}$ & $\hspace{\widthof{1}}4.88 \pm 0.04$ & $\hspace{\widthof{1}}4.57 \pm 0.01$\\
                    & CRPS & $\textbf{17.35} \pm \textbf{0.30}$ & $23.53 \pm 0.20$ & $29.22 \pm 0.08$\\

\midrule
\multirow{2}{*}{Cell count} & RMSE & $\textbf{\hspace{\widthof{1}}4.25} \pm \textbf{0.06}$ & $\hspace{\widthof{1}}4.68 \pm 0.08$ & $\hspace{\widthof{1}}4.66 \pm 0.03$\\
                    & NLL & $\textbf{\hspace{\widthof{1}}0.08} \pm \textbf{0.00}$ & $ \hspace{\widthof{1}}0.11 \pm 0.02$ & $\hspace{\widthof{1}}0.13 \pm 0.05$\\                  
\bottomrule
\end{tabular}
\end{small}
\end{table}
\begin{figure}[ht]
    \centering
    \includegraphics[width=1.0\textwidth]{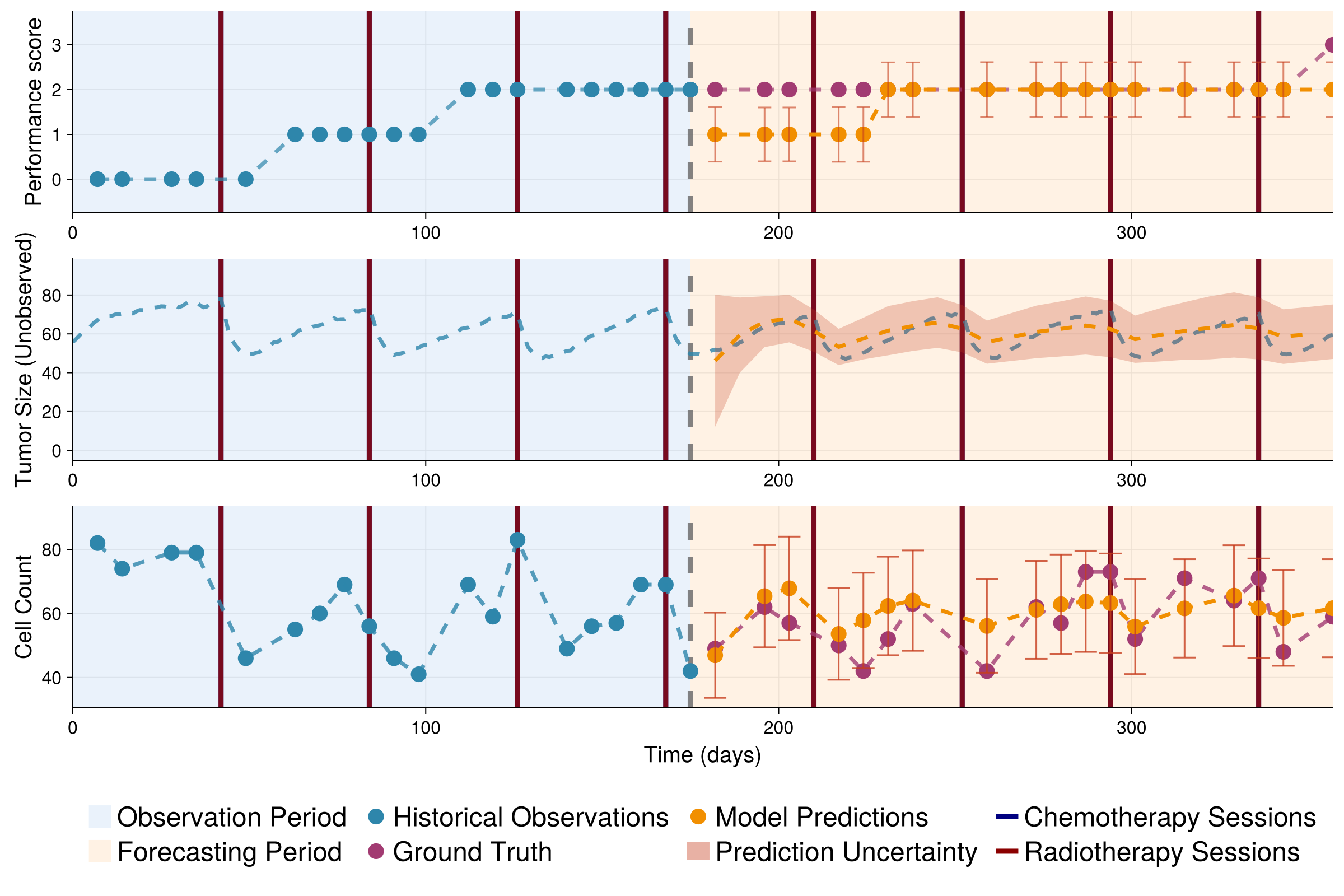}
    \caption{Illustration of the synthetic PKPD dataset with latent SDE framework predictions. The figure shows three variables over a one-year simulation period: performance score (top), tumor size in unobserved latent space (middle), and cancer cell count (bottom). Blue circles indicate historical observations during the observation period, while orange circles with error bars show model predictions during the forecasting period. Vertical red lines indicate chemotherapy sessions, and vertical brown lines indicate radiotherapy sessions. The dashed black line separates the observation period (blue shaded) from the forecasting period (orange shaded). Shaded regions around predictions represent 95\% confidence intervals, demonstrating the model's uncertainty quantification capability.}
    \label{fig:pkpd_lsde}
\end{figure}

To further assess these modeling approaches under realistic clinical conditions, we conducted systematic robustness studies examining model performance across varying levels of process noise and measurement irregularities. These experiments evaluate whether the observed performance differences persist under the challenging data characteristics commonly encountered in clinical practice, where observations are sparse, irregularly sampled, and contaminated by measurement noise.
\subsubsection{Process noise robustness}
We evaluated model performance across three process noise levels with standard deviations $\sigma \in \{0.01, 0.1, 0.5\}$ , representing low, moderate, and high noise scenarios respectively. As shown in Table \ref{tab:pkpd_performance_noise} under low noise, the latent LSTM achieved competitive performance on point prediction metrics (RMSE, Acc). However, as noise levels increased to moderate and high settings, the latent SDE demonstrated substantially better robustness, maintaining relatively stable performance while both baseline models exhibited marked deterioration. This advantage was particularly pronounced in probabilistic metrics (PE, CRPS, NLL), where the latent SDE's performance was more stable under increasing noise compared to the deterministic latent ODE and the latent LSTM. These results demonstrate that explicit stochastic modeling becomes increasingly valuable as system noise intensifies.
\begin{table}[ht]
\centering
\setlength{\tabcolsep}{2.5pt}
\caption{Model performance under varying process noise levels on the synthetic dataset. Three noise conditions are evaluated: low $(\sigma = 0.01)$, moderate $(\sigma = 0.1)$, and high $(\sigma = 0.5)$. The moderate level corresponds to the baseline experiments in Table 1. }
\label{tab:pkpd_performance_noise}
\begin{small}
\begin{tabular}{l l @{\hspace{5pt}} l c c c}
\toprule
\makecell[l]{\textbf{Target} \\ \textbf{Variable}} & \textbf{Metric} & \makecell[l]{\textbf{Noise} \\ \textbf{Level}} & \textbf{\makecell{Latent \\ SDE}} & \textbf{\makecell{Latent \\ ODE}} & \textbf{\makecell{Latent\\ LSTM}} \\
\midrule
\multirow{6}{*}{\makecell{ Performance \\ status}}
& \multirow{3}{*}{Acc} & Low & $\hspace{\widthof{1}}0.61 \pm 0.17$ & $\hspace{\widthof{1}}0.64 \pm 0.09$ & $\textbf{\hspace{\widthof{1}}0.67} \pm \textbf{0.02}$ \\
& & Moderate & $\textbf{\hspace{\widthof{1}}0.56} \pm \textbf{0.08}$ & $\hspace{\widthof{1}}0.47 \pm 0.11$ & $\hspace{\widthof{1}}0.44 \pm 0.07$ \\
& & High & $\textbf{\hspace{\widthof{1}}0.36} \pm \textbf{0.16}$ & $\hspace{\widthof{1}}0.29 \pm 0.22$ & $\hspace{\widthof{1}}0.21 \pm 0.14$ \\
\cmidrule{2-6}
& \multirow{3}{*}{PE} & Low & $\textbf{\hspace{\widthof{1}}7.01} \pm \textbf{1.10}$ & $\hspace{\widthof{1}}7.88 \pm 0.66$ & $\hspace{\widthof{1}}7.12 \pm 0.11$ \\
& & Moderate & $\textbf{\hspace{\widthof{1}}4.56} \pm \textbf{0.69}$ & $\hspace{\widthof{1}}6.43 \pm 0.47$ & $\hspace{\widthof{1}}8.56 \pm 0.21$ \\
& & High & $\textbf{\hspace{\widthof{1}}8.33} \pm \textbf{0.87}$ & $\hspace{\widthof{1}}11.04 \pm 1.39$ & $\hspace{\widthof{1}}17.23 \pm 1.55$ \\
\midrule
\multirow{6}{*}{\makecell{Tumor volume \\ (unobserved)}}
& \multirow{3}{*}{RMSE} & Low & $\hspace{\widthof{1}}4.74 \pm 0.26$ & $\hspace{\widthof{1}}4.33 \pm 0.03$ & $\textbf{\hspace{\widthof{1}}3.11} \pm \textbf{0.01}$ \\
& & Moderate & $\textbf{\hspace{\widthof{1}}4.27} \pm \textbf{0.06}$ & $\hspace{\widthof{1}}4.88 \pm 0.04$ & $\hspace{\widthof{1}}4.57 \pm 0.01$ \\
& & High & $\textbf{\hspace{\widthof{1}}7.81} \pm \textbf{0.60}$ & $\hspace{\widthof{1}}9.82 \pm 1.31$ & $\hspace{\widthof{1}}13.01 \pm 2.44$ \\
\cmidrule{2-6}
& \multirow{3}{*}{CRPS} & Low & $\hspace{\widthof{1}}17.26 \pm 1.20$ & $\textbf{\hspace{\widthof{1}}16.78} \pm \textbf{0.20}$ & $\hspace{\widthof{1}}17.22 \pm 0.16$ \\
& & Moderate & $\textbf{\hspace{\widthof{1}}17.35} \pm \textbf{0.30}$ & $\hspace{\widthof{1}}23.53 \pm 0.20$ & $\hspace{\widthof{1}}29.22 \pm 0.08$ \\
& & High & $\textbf{\hspace{\widthof{1}}22.02} \pm \textbf{0.78}$ & $\hspace{\widthof{1}}27.89 \pm 1.30$ & $\hspace{\widthof{1}}31.02 \pm 2.77$\\
\midrule
\multirow{6}{*}{Cell count}
& \multirow{3}{*}{RMSE} & Low & $\hspace{\widthof{1}}4.75 \pm 0.36$ & $\hspace{\widthof{1}}4.43 \pm 0.08$ & $\textbf{\hspace{\widthof{1}}3.66} \pm \textbf{0.07}$ \\
& & Moderate & $\textbf{\hspace{\widthof{1}}4.25} \pm \textbf{0.06}$ & $\hspace{\widthof{1}}4.68 \pm 0.08$ & $\hspace{\widthof{1}}4.66 \pm 0.03$ \\
& & High & $\textbf{\hspace{\widthof{1}}10.12} \pm \textbf{0.87}$ & $\hspace{\widthof{1}}12.02 \pm 1.44$ & $\hspace{\widthof{1}}13.77 \pm 1.95$ \\
\cmidrule{2-6}
& \multirow{3}{*}{NLL} & Low & $\hspace{\widthof{1}}0.09 \pm 0.02$ & $\hspace{\widthof{1}}0.05 \pm 0.01$ & $\textbf{\hspace{\widthof{1}}0.02} \pm \textbf{0.01}$ \\
& & Moderate & $\textbf{\hspace{\widthof{1}}0.08} \pm \textbf{0.00}$ & $\hspace{\widthof{1}}0.11 \pm 0.02$ & $\hspace{\widthof{1}}0.13 \pm 0.05$ \\
& & High & $\textbf{\hspace{\widthof{1}}0.24} \pm \textbf{0.07}$ & $\hspace{\widthof{1}}0.41 \pm 0.11$ & $\hspace{\widthof{1}}1.21 \pm 0.80$ \\
\bottomrule
\end{tabular}
\end{small}
\end{table}
\subsubsection{Data irregularities robustness}
To assess robustness to irregular sampling patterns, we systematically varied data availability by randomly removing time points from patient trajectories. We evaluated three sparsity levels: low (20\%), moderate (50\%), and high (80\%) missing observations. Each irregularity condition was evaluated independently using identical training and forecasting methods.

As presented in Table \ref{tab:pkpd_irregularity_performance_comparison}, under low missingness, the latent LSTM and latent ODE achieved competitive point prediction accuracy, benefiting from sufficient temporal resolution. However, as sparsity increased, the latent SDE demonstrated superior robustness, maintaining relatively stable performance while both baselines showed considerable degradation. This advantage was most evident at moderate sparsity, where the latent SDE outperformed baselines across nearly all metrics. Under extreme sparsity (80\% missing), the continuous-time models (SDE and ODE) both showed better resilience than the discrete-time LSTM, with the latent SDE maintaining the strongest overall performance.

These robustness evaluations demonstrate that the latent SDE framework maintains advantages under data degradation conditions—high process noise and irregular sampling—that are common in clinical practice. Having established these properties on controlled synthetic data, we next evaluate performance on real-world physiological time series from ICU patients.

\begin{table}[ht]
\centering
\setlength{\tabcolsep}{0.8pt}
\caption{Model performance under varying data irregularity levels on the synthetic dataset. Three sparsity conditions are evaluated: low (20\% missing), moderate (50\% missing), and high (80\% missing). The moderate level corresponds to the baseline experiments in Table 1.}
\label{tab:pkpd_irregularity_performance_comparison}
\begin{small}
\begin{tabular}{l l l c c c}
\toprule
\makecell{\textbf{Target} \\ \textbf{Variable}} & \textbf{Metric} & \textbf{ \makecell{Missingness \\ Level}} & \textbf{\makecell{Latent \\ SDE}} & \textbf{\makecell{Latent \\ ODE}} & \textbf{\makecell{Latent\\ LSTM}} \\
\midrule
\multirow{6}{*}{\makecell{ Performance \\ status}}
& \multirow{3}{*}{Acc} & Low (20\%) & $\hspace{\widthof{1}}0.62 \pm 0.12$ & $\hspace{\widthof{1}}0.67 \pm 0.08$ & $\textbf{ \hspace{\widthof{1}}0.73} \pm \textbf{0.03}$ \\
& & Moderate (50\%) & $\textbf{\hspace{\widthof{1}}0.56} \pm \textbf{0.08}$ & $\hspace{\widthof{1}}0.47 \pm 0.11$ & $\hspace{\widthof{1}}0.44 \pm 0.07$ \\
& & High (80\%)& $\hspace{\widthof{1}}0.42 \pm 0.13$ & $\textbf{\hspace{\widthof{1}}0.43} \pm \textbf{0.21}$ & $\hspace{\widthof{1}}0.32 \pm 0.15$ \\
\cmidrule{2-6}
& \multirow{3}{*}{PE} & Low (20\%) & $\hspace{\widthof{1}}3.22 \pm 0.12$ & $\hspace{\widthof{1}}3.43 \pm 0.31$ & $\textbf{2.55} \pm \textbf{\hspace{\widthof{1}}0.20}$ \\
& & Moderate (50\%) & $\textbf{\hspace{\widthof{1}}4.56} \pm \textbf{0.69}$ & $\hspace{\widthof{1}}6.43 \pm 0.47$ & $\hspace{\widthof{1}}8.56 \pm 0.21$ \\
& & High (80\%) & $\textbf{\hspace{\widthof{1}}8.87} \pm \textbf{1.44}$ & $\hspace{\widthof{1}}9.01 \pm 1.02$ & $\hspace{\widthof{1}}13.04 \pm 0.97$ \\
\midrule
\multirow{6}{*}{\makecell{Tumor volume \\ (unobserved)}}
& \multirow{3}{*}{RMSE} & Low (20\%) & $\hspace{\widthof{1}}5.77 \pm 0.26$ & $\hspace{\widthof{1}}3.73 \pm 0.14$ & $\textbf{\hspace{\widthof{1}}3.22} \pm \textbf{0.09}$ \\
& & Moderate (50\%) & $\textbf{\hspace{\widthof{1}}4.27} \pm \textbf{0.06}$ & $\hspace{\widthof{1}}4.88 \pm 0.04$ & $\hspace{\widthof{1}}4.57 \pm 0.01$ \\
& & High (80\%) & $\textbf{\hspace{\widthof{1}}4.14} \pm \textbf{0.22}$ & $\hspace{\widthof{1}}5.13 \pm 0.22$ & $\hspace{\widthof{1}}6.98 \pm 0.89$ \\
\cmidrule{2-6}
& \multirow{3}{*}{CRPS} & Low (20\%) & $\hspace{\widthof{1}}19.31 \pm 0.30$ & $\textbf{\hspace{\widthof{1}}18.91} \pm \textbf{0.11}$ & $\hspace{\widthof{1}}18.98 \pm 0.08$ \\
& & Moderate (50\%) & $\textbf{\hspace{\widthof{1}}17.35} \pm \textbf{0.30}$ & $\hspace{\widthof{1}}23.53 \pm 0.20$ & $\hspace{\widthof{1}}29.22 \pm 0.08$ \\
& & High (80\%) & $\hspace{\widthof{1}}22.26 \pm 0.89 $ & $\textbf{\hspace{\widthof{1}}19.74} \pm \textbf{0.44}$ & $\hspace{\widthof{1}}28.12 \pm 1.33$ \\
\midrule
\multirow{6}{*}{Cell count}
& \multirow{3}{*}{RMSE} & Low (20\%) & $\hspace{\widthof{1}}5.98 \pm 0.76$ & $\hspace{\widthof{1}}3.78 \pm 0.08$ & $\textbf{\hspace{\widthof{1}}3.01} \pm \textbf{0.19}$ \\
& & Moderate (50\%) & $\textbf{\hspace{\widthof{1}}4.25} \pm \textbf{0.06}$ & $\hspace{\widthof{1}}4.68 \pm 0.08$ & $\hspace{\widthof{1}}4.66 \pm 0.03$ \\
& &  High (80\%) & $\textbf{\hspace{\widthof{1}}4.77} \pm \textbf{0.27}$ & $\hspace{\widthof{1}}5.01 \pm 0.45$ & $\hspace{\widthof{1}}7.02 \pm 0.34$ \\
\cmidrule{2-6}
& \multirow{3}{*}{NLL} & Low (20\%) & $\textbf{\hspace{\widthof{1}}0.13} \pm \textbf{0.01}$ & $\hspace{\widthof{1}}0.14 \pm 0.01$ & $\hspace{\widthof{1}}0.13 \pm 0.05$ \\
& & Moderate (50\%)& $\textbf{\hspace{\widthof{1}}0.08} \pm \textbf{0.00}$ & $\hspace{\widthof{1}}0.11 \pm 0.02$ & $\hspace{\widthof{1}}0.13 \pm 0.05$ \\
& & High (80\%) & $\textbf{\hspace{\widthof{1}}0.17} \pm \textbf{0.00}$ & $\hspace{\widthof{1}}0.18 \pm 0.08$ & $\hspace{\widthof{1}}0.25 \pm 0.00$ \\
\bottomrule
\end{tabular}
\end{small}
\end{table}
\subsection{Empirical dataset}
In this task, we adopted a probabilistic forecasting approach where models predict Gaussian distributions $\mathcal{N}(\mu,\sigma^2))$ for each time point, rather than single-point predictions. This approach enables better uncertainty quantification, which is essential in high-stakes clinical decision-making. Moreover, we incorporated patient baseline covariates, including age, gender, and height into the observational data. These covariates were selected due to their established clinical relevance in predicting patient conditions within the ICU. 

\begin{figure}[ht]
    \centering
    \includegraphics[width=1.0\textwidth]{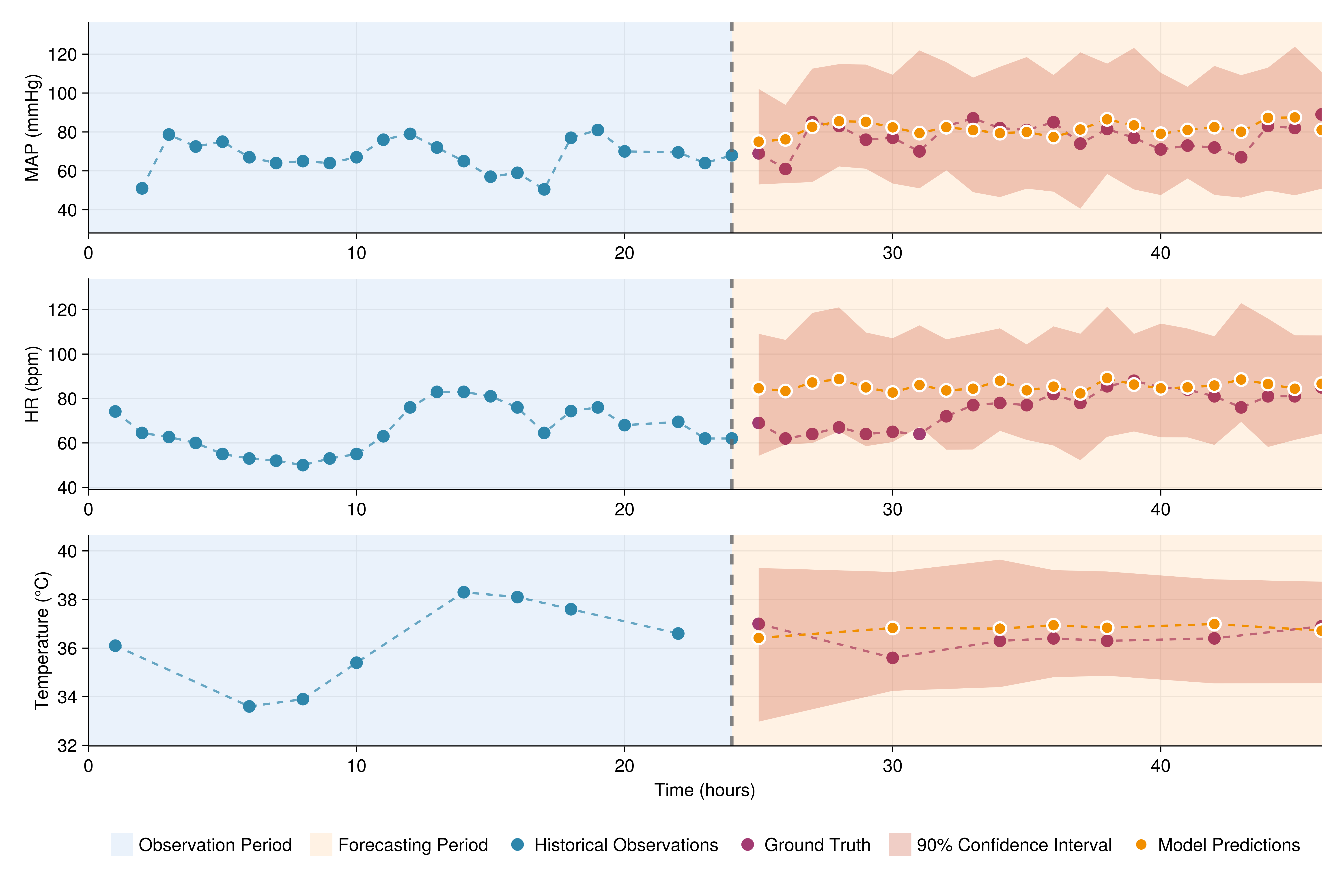}
    \caption{Sample ICU patient trajectories from the PhysioNet 2012 dataset. The figure shows three vital signs over 48 hours: heart rate (top), mean arterial pressure (middle), and body temperature (bottom). Blue circles represent observed measurements during the first 24 hours (observation period, blue shaded region), while orange circles show model predictions for the subsequent 24 hours (forecasting period, orange shaded region). Orange dashed lines indicate predicted mean values, and shaded regions represent 95\% confidence intervals. The vertical dashed line separates the observation and forecasting periods, demonstrating the model's ability to forecast vital signs with calibrated uncertainty estimates.}
    \label{fig:icu_sample_trajectory}
\end{figure}

Similar to the synthetic data, we considered 50\% of the patient data (the first 24 hours of each patient's stay) as the observation period and aimed to predict the subsequent 50\% data (next 24 hours). Figure \ref{fig:icu_sample_trajectory} illustrates a sample patient trajectory from the ICU dataset, demonstrating the observation and prediction periods, alongside the latent SDE model predictions. The figure shows how the model captures the general trends in vital signs while providing uncertainty bands that reflect the inherent variability in physiological measurements.

Table~\ref{tab:models_performance_icu} presents the forecasting performance of all models on the ICU dataset for the selected three physiological variables: mean arterial pressure (MAP), heart rate (HR), and body temperature (BT). The latent SDE achieved the best overall performance, producing more accurate point predictions (lower RMSE) and better-calibrated uncertainty estimates (lower CRPS) for heart rate and body temperature. For mean arterial pressure, all models performed similar, with the latent LSTM showing marginally better performance. This similarity in MAP prediction performance may reflect the relatively stable nature of mean arterial pressure in ICU settings, where it is often tightly controlled through medical interventions, reducing the benefit of explicit stochasticity modeling. 
\begin{table}[ht]
\centering
\caption{Performance comparison on PhysioNet 2012 ICU dataset for 24-hour vital sign forecasting. Three variables are predicted: mean arterial pressure (MAP), heart rate (HR), and body temperature (BT). Lower values indicate better performance for both RMSE and CRPS.}
\label{tab:models_performance_icu}
\setlength{\tabcolsep}{4pt}
\begin{small}
\begin{tabular}{l l c c c}
\toprule
 \textbf{\makecell{Target \\Variable}} &  \textbf{Metrics} &  \makecell{\textbf{Latent} \\ \textbf{SDE}} &  \makecell{\textbf{Latent} \\ \textbf{ODE}} & \makecell{\textbf{Latent} \\ \textbf{LSTM}} \\
\midrule
\multirow{2}{*}{MAP} & RMSE & $\hspace{\widthof{1}}0.5286 \pm 0.0061$ & $\hspace{\widthof{1}}0.5297 \pm 0.0054$ & $\textbf{\hspace{\widthof{1}}0.5257} \pm \textbf{0.0050}$ \\
                    & CRPS & $\hspace{\widthof{1}}0.2031 \pm 0.0029$ & $\hspace{\widthof{1}}0.2071 \pm 0.0054$ & $\textbf{\hspace{\widthof{1}}0.2009} \pm \textbf{0.0007}$\\
\midrule
\multirow{2}{*}{HR} & RMSE & $\textbf{\hspace{\widthof{1}}0.8193} \pm \textbf{0.0134}$ & $\hspace{\widthof{1}}0.8261 \pm 0.0014$ & $\hspace{\widthof{1}}0.8197 \pm 0.0120$\\
                    & CRPS & $\textbf{\hspace{\widthof{1}}0.4779} \pm \textbf{0.0139}$ & $\hspace{\widthof{1}}0.4890 \pm 0.0141$ & $\hspace{\widthof{1}}0.4793 \pm 0.0119$\\

\midrule
\multirow{2}{*}{BT} & RMSE &$ \textbf{\hspace{\widthof{1}}0.3461}  \pm \textbf{0.0027} $ & $\hspace{\widthof{1}}0.3501 \pm 0.0065$ & $\hspace{\widthof{1}}0.3593 \pm   0.0045$\\
                    & CRPS &$\textbf{\hspace{\widthof{1}}0.0910} \pm \textbf{0.0012}$  & $\hspace{\widthof{1}}0.0931 \pm 0.0029$ &$\hspace{\widthof{1}}0.0975 \pm 0.0006$  \\                  
\bottomrule
\end{tabular}
\end{small}
\end{table}
These results demonstrate that the latent SDE framework's advantages observed in synthetic experiments extend to real-world clinical data, validating its practical utility for modeling the complex, stochastic dynamics of ICU patient monitoring. The framework's ability to provide well-calibrated uncertainty estimates alongside accurate predictions makes it particularly valuable for clinical applications where understanding the confidence of forecasts is as important as the forecasts themselves.

\section{Discussion}
Our proposed latent stochastic differential equation (SDE) framework demonstrates significant advances in modeling clinical time series data. The model outperformed latent ODE and latent LSTM baselines across the synthetic and real-world ICU datasets (Tables~\ref{tab:model_performance_pkpd} and~\ref{tab:models_performance_icu}), while maintaining robust performance under challenging conditions, such as high irregularities and  varying noise levels (Tables~\ref{tab:pkpd_performance_noise} and~\ref{tab:pkpd_irregularity_performance_comparison}). This robustness underscores the framework’s potential for real-world clinical applications where data is often limited, irregular and noisy.

The latent SDE framework’s success stems from its stochastic and continuous-time formulation. Unlike deterministic ODEs, SDEs capture inherent randomness in patient trajectories through the diffusion term, enabling more accurate modeling of complex physiological dynamics (Fig. \ref{fig:pkpd_lsde}). The stochastic component allows the model to represent the natural variability in how different patients respond to the same treatment, accounting for factors such as individual metabolism, comorbidities, and unmeasured confounders. Additionally, the continuous-time nature of SDEs naturally handles irregular sampling, a common challenge in clinical data. In contrast, the latent LSTM model struggled with irregular data, long-term dependencies, and uncertainty quantification mainly because of its discrete time architecture, limited capacity to model stochastic processes, posterior collapse in variational inference and sensitivity to vanishing gradients. While latent ODEs performed better than LSTMs, their deterministic nature limited their ability to model stochasticity, leading to increased errors in noisy conditions.

The ability to quantify predictive uncertainty is a key strength of the latent SDE approach. For instance, lower CRPS values indicate well-calibrated uncertainty estimates, which can guide clinicians in high-stakes decisions, such as prioritizing early interventions for ICU patients with unstable vital signs (e.g., erratic heart rate patterns signaling potential sepsis). By providing both expected trajectories and their associated uncertainties, the model supports risk-aware decision-making, enhancing its utility in personalized medicine, and treatment optimization. 

Despite these strengths, the framework has limitations that should be acknowledged. Training SDEs is difficult and computationally intensive compared to ODEs or LSTMs, which may pose challenges in resource-constrained settings or when rapid model development is required. Additionally, the model assumes Markovian latent dynamics, which may not fully capture non-stationary processes in some clinical scenarios. The reliance on synthetic PKPD data, while useful for controlled experiments, may not fully reflect real-world complexities, such as coexisting health conditions. Furthermore, our real-world evaluation focused on relatively short-term forecasting (24 hours) in ICU settings; performance over longer horizons or in other clinical contexts remains to be established.

Future work could enhance the framework by integrating domain knowledge, such as expert-designed ODEs, to improve interpretability and align with clinical insights~\citep{qian2021integrating}. Additionally, extending the model to incorporate multi-modal data, such as imaging or genomics, could enable more comprehensive patient modeling~\citep{bazgir2023integration, vanguri2022multimodal}. Exploring real-time forecasting applications, such as online prediction of ICU vital signs, could further support dynamic treatment planning, enhancing clinical utility~\citep{zhang2025intelligent, lim2024real}.This would require developing efficient online inference methods that update predictions as new measurements arrive, enhancing clinical utility for time-sensitive decisions.

Another promising direction is developing methods for treatment policy optimization using the learned SDE model. Given the model's ability to simulate counterfactual outcomes under different treatment strategies, it could be combined with reinforcement learning or optimal control methods to recommend personalized treatment plans. However, such applications would require careful validation and consideration of safety constraints before clinical deployment.

In conclusion, our latent SDE framework offers a robust solution for modeling clinical time series, providing accurate and uncertainty-aware predictions that may advance personalized medicine. Its ability to handle irregular, noisy data and quantify uncertainty makes it a promising tool for tasks such as disease progression modeling, treatment effect estimation, and treatment optimization, paving the way for more precise and reliable clinical decision-making. As healthcare increasingly adopts data-driven approaches, frameworks that combine predictive accuracy with reliable uncertainty quantification will be essential for safe and effective clinical AI systems.

\section*{Acknowledgments}
This study is supported by the Radboud Healthy Data program, which is partially funded by the Reinier Post Foundation.
\bibliographystyle{elsarticle-num-names} 
\bibliography{references}

\appendix
\label{appndx}
\section{Data}
\label{appndx:data}
\subsection{Synthetic dataset}
\label{appndx:synthetic_dataset}
In the following, we provide more details on the synthetic dataset generation using the lung cancer pharmacokinetic-pharmacodynamic (PKPD) model. This model is widely used in the literature to study the effects of chemotherapy and radiotherapy treatments on tumor growth~\citep{kacprzyk2024ode}. However, to achieve a more realistic simulation, we have expanded the model beyond tumor size to incorporate both immune system response and patient overall health status. These additions allow the model to capture not only the direct impact of treatments on the tumor but also their effects on immune system dynamics and the patient’s physiological condition, reflecting the complex interplay of factors that influence treatment outcomes in real clinical settings.

In this model, tumor size $x(t)$ evolves according to Equation~\eqref{eq:pkpd_main_eq}. This differential equation describes tumor volume $x(t)$ following a logistic growth model with carrying capacity $K$, where chemotherapy ($\beta_cc(t)$) and radiotherapy ($\alpha_rd(t)+\beta_rd(t)^2$) effects reduce the growth rate. The noise term $e(t)$ captures random fluctuations, and the entire expression is scaled by $x(t)$, reflecting growth proportional to the current tumor size.
Tumor growth dynamics are given by:
\begin{align}
\dd x(t) &= \left( \rho \log \left(\frac{K}{x(t)} \right) - \beta_c c(t) - (\alpha_r d(t) + \beta_r d(t)^2)  \right) x(t) \dd t + \sigma x(t) \dd W
\label{eq:pkpd_main_eq}
\end{align}
where $c(t)$, and $d(t)$ are the concentrations of chemotherapy and radiotherapy drugs at time $t$,  each described by its own set of differential equations as following: 
\begin{align}
\frac{\dd c(t)}{\dd t} &= -\phi_c \, c(t) + u_c(t) 
\label{eq:chemo_eq} \\
\frac{\dd d(t)}{\dd t} &= -\phi_d \, d(t) + u_d(t)
\label{eq:radio_eq}
\end{align}
where $u_c(t) \in \{0,1\}$ and $u_d(t)\in \{0,1\}$ are chemotherapy and treatment administration respectively, while $\phi_c$ and $\phi_d$ represent half-life time of the corresponding drug. These equations model the standard pharmacokinetic principle of exponential drug clearance with periodic administration.

The dynamics of the immune system response \( I(t) \) are modeled by:
\begin{align}
\frac{\dd I(t)}{\dd t} &= \delta \left(1 - \frac{I(t)}{I_{\text{max}}}\right) I(t) - \beta_I c(t) - \alpha_I d(t) + 
\theta_I \, \frac{I_{\text{max}} - I(t)}{1 + \lambda_I x(t)} - \omega_I I(t) 
\label{eq:immune_eq}
\end{align}
showing that an increase in tumor volume \( x(t) \) reduces \( I(t) \) by inhibiting immune stimulation. Additionally, higher concentrations of chemotherapy \( c(t) \) and radiotherapy \( d(t) \) suppress \( I(t) \), reflecting their immunosuppressive effects. The equation also includes logistic growth, stimulation toward a maximum capacity \( I_{\text{max}} \), and natural decay, balancing the immune response between treatments and tumor burden.
 
The patient's overall health dynamics are defined by
\begin{align}
\frac{\dd S(t)}{\dd t} &= \frac{\theta_S (1 - S(t))}{1 + \lambda_S x(t)} - \gamma_I \left(\frac{I(t)}{I_{\text{max}}} - 1\right)^2 \,.
\label{eq:health_eq}
\end{align} 
Consequently, an increase in the tumor volume degrades patient health status, and any deviation from the optimal immune system response also has a detrimental effect on the patient health status. Therefore, optimizing chemotherapy and radiotherapy treatments to reduce tumor volume, while considering their effects on the immune system and health status, is crucial.

To account for between-patient variability, we represented each parameter as a distribution rather than a single fixed value. Moreover, we considered five baseline covariates for each patient: gender, age, height, weight, and tumor type. These covariates affect the model parameters, thereby influencing disease progression, patient treatment response, and health status. For instance, older patients are more susceptible to faster tumor growth and slower health recovery. Furthermore, these covariates were used to derive more complex, patient-specific covariates. Specifically, body mass index (BMI) and body surface area (BSA), calculated from height and weight, were included due to their relevance to cancer progression and treatment.
\begin{table}[h!]
\centering
\caption{Parameters and baseline covariates for the PKPD model of lung cancer. Model parameters follow Normal$(\mu, \sigma)$ distributions to create patient-specific heterogeneity in treatment responses.}
\label{tab:cancer_pkpd_parameters}
\begin{scriptsize}
\setlength{\tabcolsep}{4pt}  
\begin{tabularx}{\textwidth}{>{\raggedright\arraybackslash}p{1.8cm}>{\raggedright\arraybackslash}p{3.5cm}cc>{\raggedright\arraybackslash}X}
\toprule
\textbf{Model} & \textbf{Variable} & \textbf{Parameter} & \textbf{Distribution} & \textbf{Parameter Value $(\mu, \sigma)$} \\
\midrule
\multirow{6}{*}[-10pt]{\makecell{Tumor \\ growth}} 
& Growth rate & $\rho$ & Normal & $8.00 \times 10^{-2}$, $1.00 \times 10^{-3}$ \\
& Carrying capacity & $K$ & Normal & $100.0$, $10.0$ \\
& Radiotherapy linear effect & $\alpha_r$ & Normal & $0.1$, $0.05$\\
& Radiotherapy quadratic effect & $\beta_r$ & Normal & $0.1$, $0.05$ \\
& Chemotherapy linear effect & $\alpha_c$ & Normal & $0.1$, $0.05$ \\
& Chemotherapy quadratic effect & $\beta_c$ & Normal & $0.1$, $0.05$ \\
\midrule
Radiotherapy & Radiotherapy drug half-life time & $\phi_r$ & Normal & $0.1$, $0.05$ \\
\midrule
Chemotherapy & Chemotherapy drug half-life time & $\phi_c$ & Normal & $0.1$, $0.05$ \\ 
\midrule
\multirow{7}{*}[-15pt]{\makecell{Immune \\System}} 
& Immune cell growth rate & $\delta$ & Normal & $0.013$, $0.005$ \\
& Max immune response & $I_{\text{max}}$ & Normal & $0.95$, $0.4$ \\
& Chemotherapy effect on immune & $\beta_I$ & Normal & $0.1$, $0.05$ \\
& Radiotherapy effect on immune & $\alpha_I$ & Normal & $0.1$, $0.05$ \\
& Immune stimulation by tumor & $\theta_I$ & Normal & $0.08$, $0.04$ \\
& Immune suppression by tumor & $\lambda_I$ & Normal & $0.005$, $0.002$ \\
& Immune decay rate & $\omega_I$ & Normal & $0.15$, $0.05$ \\
\midrule
\multirow{3}{=}{Overall Health} 
& Immune effect on health & $\gamma_I$ & Normal & $8.00 \times 10^{-3}$, $5 \times 10^{-3}$ \\
& Health recovery rate & $\theta_S$ & Normal & $100.0$, $10.0$ \\
& Tumor impact on health & $\lambda_S$ & Normal & $200.0$, $20.0$ \\
\midrule
\multirow{5}{=}{Baseline Covariates} 
& Age (years) & & Constant & 80--120 \\
& Weight (kg) & & Constant & 70--150 \\
& Height (cm) & & Constant & 100--200 \\
& Gender & & Binary & Male, Female \\
& Tumor Type & & Binary & Small cell, Non-small cell \\
\bottomrule
\end{tabularx}
\end{scriptsize}
\end{table}
We also considered indirect observations of the system states. Assuming that tumor size $x(t)$ is observed indirectly via cancer cell count, modeled as a Poisson-distributed observation according to:\begin{align}
\label{eq:poisson_obs}
y(t) &\sim \text{Poisson}(x(t)) \,.
\end{align}
Additionally, the health status is observed through performance scores like the Eastern Cooperative Oncology Group (ECOG) performance status~\citep{oken1982ecog}, which are standardized tools used in oncology to assess a patient's level of functioning and their ability to perform daily activities. These scores, which typically range from 0 (fully active) to 5 (deceased), are crucial for determining treatment suitability, prognosis, and a patient's overall quality of life.
To classify the patient health status into performance scores, we defined a function that maps the continuous health value $S \in [0,1]$ to discrete performance scores categories ranging from 0 (fully active) to 5 (deceased). This approach provides a smooth and interpretable conversion of a continuous health status into a clinically relevant ordinal scale (Table~\ref{tab:performace_score}).

\begin{table}[h!]
    \centering
        \caption{Eastern Cooperative Oncology Group (ECOG) Performance Score~\citep{oken1982ecog}.}
\begin{small}
\begin{tabularx}{\textwidth}{cX} 
        \toprule
        \makecell{\textbf{ECOG} \\ \textbf{Performance Score}} & \textbf{Description} \\
        \midrule
        0 & Fully active, able to carry on all pre-disease activities without restriction. \\
        1 & Restricted in physically strenuous activity but ambulatory and able to carry out light or sedentary work (e.g., office work, light housework). \\
        2 & Ambulatory and capable of all self-care but unable to carry out any work activities; up and about more than 50\% of waking hours. \\
        3 & Capable of only limited self-care; confined to bed or chair more than 50\% of waking hours. \\
        4 & Completely disabled; cannot carry on any self-care; totally confined to bed or chair. \\
        5 & Deceased. \\
        \bottomrule
    \end{tabularx}
\end{small}    \label{tab:performace_score}
\end{table}

Ultimately, we randomly dropped some time points from each trajectory to simulate irregularly sampled measurements (observations).
We utilized the explained models to generate synthetic data, simulating the effects of treatments on cancer progression and patient health. The simulation covered a one-year period, with observations recorded weekly and treatments applied at individualized intervals, such as every 2 weeks for one patient or every 4 to 5 weeks for another patient. This approach allowed us to explore the dynamic responses to therapy while accounting for the temporal structure of the clinical interventions, aligning closely with real-world settings.

\subsection{Empirical dataset}
\label{appndx:empirical_dataset}
The PhysioNet Computing in Cardiology Challenge 2012 dataset is a widely recognized and benchmarked collection developed specifically for advancing predictive modeling and analytics in clinical settings. Comprising extensive patient data collected from multiple hospitals, this dataset includes vital signs, clinical measurements, laboratory test results, and demographic information for a diverse group of ICU patients. Each patient's data consists of multivariate time-series records captured during their ICU stay, providing a detailed temporal representation of their clinical condition. The dataset includes records from approximately 12,000 ICU admissions with minimum of 48 hours of stay. It contains a total of 42 variables, with 6 general descriptive variables and 36 time-dependent variables measured at least once per patient stay. Additionally, five important outcome-related descriptors such as survival status, total length of ICU stay, and occurrence of in-hospital death are recorded for each patient. A detailed summary of these variables is provided in Tables~\ref{tab:icu_timeseries_variables}, ~\ref{tab:icu_general_descriptors}, and ~\ref{tab:icu_outcome_descriptors}.

This dataset was initially created for a global competition focused on predicting mortality risk among patients admitted to the ICU. However, in this work, we utilized it to predict the time series values rather than mortality prediction. Nonetheless, not all time series variables in the dataset are suitable for this application, mainly because some variables are not measured frequent enough. For example, many laboratory test results are often recorded only two or three times per patient, which restricts their effectiveness as reliable time series data. Accordingly, we selected mean atrial blood pressure (MAP), heart rate (HR), and body temperature (BT) as our target variables to predict. Additionally, this dataset features significant irregularities in measurement intervals, as most variables are recorded at irregular time points. Moreover, the lengths of time series for individual variables vary considerably. These unique characteristics make the dataset particularly valuable for exploring and assessing the proposed model's ability to handle irregular and asynchronous clinical data measurements.

\begin{table}[h!]
    \centering
    \caption{Time series variables in the PhysioNet/Computing in Cardiology Challenge 2012 dataset.}
    \begin{small}
    \begin{tabularx}{\textwidth}{lX}
        \toprule
        \textbf{Variable} & \textbf{Description} \\
        \midrule
        Albumin           & Albumin (g/dL) \\
        ALP               & Alkaline phosphatase (IU/L) \\
        ALT               & Alanine transaminase (IU/L) \\
        AST               & Aspartate transaminase (IU/L) \\
        Bilirubin         & Bilirubin (mg/dL) \\
        BUN               & Blood urea nitrogen (mg/dL) \\
        Cholesterol       & Cholesterol (mg/dL) \\
        Creatinine        & Serum creatinine (mg/dL) \\
        HCO3              & Serum bicarbonate (mmol/L) \\
        HCT               & Hematocrit (\%) \\
        HR                & Heart rate (beats per minute) \\
        K                 & Serum potassium (mEq/L) \\
        Lactate           & Lactate concentration (mmol/L) \\
        Mg                & Serum magnesium (mEq/L) \\
        MAP               & Invasive mean arterial blood pressure (mmHg) \\
        MechVent          & Mechanical ventilation respiration (binary) \\
        NI DiasABP        & Non-invasive diastolic arterial blood pressure (mmHg) \\
        NI MeanABP        & Non-invasive mean arterial blood pressure (mmHg) \\
        NI SysABP         & Non-invasive systolic arterial blood pressure (mmHg) \\
        PaCO2             & Partial pressure of arterial CO\textsubscript{2} (mmHg) \\
        PaO2              & Partial pressure of arterial O\textsubscript{2} (mmHg) \\
        Platelets         & Platelet count (cells/nL) \\
        RespRate          & Respiration rate (breaths per minute) \\
        SaO2              & Oxygen saturation in hemoglobin (\%) \\
        SpO2              & Oxygen saturation (\%) \\
        Temp              & Body temperature (°C) \\
        Urine             & Urine output (mL) \\
        WBC               & White blood cell count (cells/nL) \\
        Weight            & Patient weight (kg) \\
        \bottomrule
    \end{tabularx}
\end{small}    \label{tab:icu_timeseries_variables}
\end{table}

\begin{table}[h!]
    \centering
    \caption{General descriptors (baseline characteristics) in the PhysioNet/Computing in Cardiology Challenge 2012 dataset, collected at the time of ICU admission}
    \small\begin{tabularx}{\textwidth}{lX}
        \toprule
        \textbf{Variable} & \textbf{Description} \\
        \midrule
        RecordID & Unique integer for each ICU stay \\
        Age      & Age (years)\\
        Gender   & Gender (0: female, 1: male) \\
        Height   & Height (cm)\\
        ICUType  & ICU type (1: Coronary Care Unit, 2: Cardiac Surgery Recovery Unit, 3: Medical ICU, 4: Surgical ICU)  \\
        Weight   & Weight (kg) \\
        \bottomrule
    \end{tabularx}
    \label{tab:icu_general_descriptors}
\end{table}

\begin{table}[h!]
    \centering
    \caption{Outcome-related descriptors in the PhysioNet/Computing in Cardiology Challenge 2012 dataset}
    \small\begin{tabularx}{\textwidth}{lX}
        \toprule
        \textbf{Variable} & \textbf{Description} \\
        \midrule
        RecordID          & Unique integer for each ICU stay \\
        SAPS-I score      & Simplified Acute Physiology Score~\citep{le1984simplified} \\
        SOFA score        & Sequential Organ Failure Assessment score~\citep{ferreira2001serial} \\
        Length of stay    & Duration of ICU stays (days) \\
        Survival days     & Survival duration (days) \\
        In-hospital death & In-hospital death status (0: survivor, 1: died in-hospital) \\
        \bottomrule
    \end{tabularx}
    \label{tab:icu_outcome_descriptors}
\end{table}

\section{Evaluation metrics}
We employ different evaluation metrics tailored to the specific characteristics and distributions of each target variable in our datasets.
These metrics assess both point prediction accuracy and the quality of uncertainty quantification, which are both essential for evaluating probabilistic forecasting models.

\subsection{Synthetic dataset metrics} 
For the discrete performance status variable (ECOG scores 0-5), we use two complementary metrics. First, accuracy (Acc) as the proportion of correctly classified performance status predictions, providing a straightforward measure of classification performance. Second, predictive entropy (PE), which measures the uncertainty in categorical predictions. For a predicted distribution $p(y \mid x)$, predictive entropy is defined as:
\begin{align}
  \text{PE}(p) = -\sum_{y} p(y \mid x) \log p(y|x)  
\end{align}
where lower entropy values indicate more confident and accurate predictions, while higher entropy reflects greater uncertainty in the categorical distribution.

Since tumor volume represents a continuous variable we evaluate it using two other complementary metrics. First, the root mean square error (RMSE), which measures the average magnitude of prediction errors for the continuous valued variable (tumor volume). Second, the continuous ranked probability score (CRPS), which evaluates the quality of probabilistic forecasts by comparing the predicted cumulative distribution function to the empirical distribution of observed outcomes. CRPS rewards both accurate point predictions and well-calibrated uncertainty estimates. For a Gaussian predictive distribution with mean $\mu$, standard deviation $\sigma$, and observation $y$:
    \begin{align}
            \text{CRPS}(\mu, \sigma, y) = \sigma \left[ \frac{1}{\sqrt{\pi}} - 2\phi\left( \frac{y - \mu}{\sigma} \right) - \frac{y - \mu}{\sigma} \left( 2\Phi\left( \frac{y - \mu}{\sigma} \right) - 1 \right) \right]
    \end{align}
where $\phi$ and $\Phi$ denote the standard normal PDF and CDF, respectively. Lower CRPS values indicate better-calibrated and sharper probabilistic predictions (the predicted distribution is both accurate and confident).

For the Poisson-distributed cell count observations, we again use the RMSE to assess point prediction accuracy for the discrete count values. Additionally, we use the negative log-likelihood (NLL), which evaluates how well the predicted Poisson rate parameter $\lambda$ explains the observed count data. For observed counts $\{y_1, y_2, \ldots, y_N\}$ and predicted rates $\{\lambda_1, \lambda_2, \ldots, \lambda_N\}$, NLL can be calculated by:
\begin{align}
    \text{NLL} = \sum_{i=1}^{N} (\lambda_i - y_i \log \lambda_i + \log(y_i!))
\end{align}
where lower NLL values indicate better alignment between the model's distributional assumptions and observed count data.

\subsection{Empirical dataset metrics}
For the ICU dataset, as discussed above, we forecast three continuous physiological variables: heart rate (HR), invasive mean arterial blood pressure (MAP), and body temperature (BT). Therefore similar to the tumor volume in the synthetic dataset, we use RMSE and CRPS for evaluating models performance. 

These metrics collectively assess the model's ability to handle mixed data types (continuous, discrete, count), capture uncertainty across different variable characteristics, and perform both point prediction and probabilistic forecasting tasks under the partially observed, and noisy conditions.

\section{Implementation details}

In the following, we summarize the implementation details of our proposed latent SDE framework and baseline models. 
All experiments were conducted on a MacBook Pro M2 with 16 GB of RAM. We implemented the models using Julia 1.10.5, leveraging Lux for neural network components, DifferentialEquations and SciMLSensitivity for differential equation solvers and sensitivity analysis.
We used the AdamW optimizer with an exponential learning rate scheduler to fine-tune the training process. To improve convergence, we also applied a cyclic annealing method to the Kullback-Leibler (KL) divergence loss. We use a linear schedule with four cycles, where the value increases linearly from 0 to 10 during the first half of each cycle and remains constant during the second half, before resetting.This cyclical pattern allows the model to initially focus on reconstruction before gradually incorporating the KL regularization.

All models shared the same encoder and decoder architecture to ensure fair comparison. Each model has two different encoders: one encoder for observations, and another one for the control inputs (treatments). The observation encoder was an RNN with 2 layers with 256 neurons, designed to capture temporal dependencies in the observation data. This encoder serves two objectives: (i) estimating mean ($\mu$), and variance ($\sigma$) of the latent initial conditions distributions $\mathcal{Q}_0$, (ii) providing required context for the augmented SDE during latent SDE training. 
The control inputs encoder maps the external inputs to the latent space to be used in evolving latent dynamics. Since in both datasets, the control inputs are low dimensional, we only used an identity encoder for them. In the latent ODE and SDE models, we also interpolate these control signals to provide continuous-time inputs at arbitrary time points, enabling the model to evaluate treatment effects at any moment during trajectory evolution.

The decoder consisted of a single-layer MLP with a number of neurons equal to the output size, facilitating the reconstruction of the target variables. We limited the decoder's capacity to prevent posterior collapse in the latent models. This approach forces the latent models to learn the system's underlying dynamics rather than relying on the decoder to handle all complexity.

Similarly in the latent space, all models shared the same number of parameters. For the latent SDE, we modeled the drift and diffusion using two-layer multilayer perceptron (MLP) networks, each with a size of 64. Likewise, the latent ODE vector field was parameterized by a two MLP layers with 64 neurons. Finally, the latent LSTM also used two stacked LSTM layers, each with 64 neurons. For the latent SDE model, we employed the Euler-Maruyama solver with a step size of 0.01. The latent ODE model used the Euler solver with the same step size. These relatively small step sizes ensure accurate numerical integration, particularly important for capturing rapid changes in patient state.

\end{document}